\documentclass{article}

\usepackage{arxiv}

\usepackage[utf8]{inputenc} 
\usepackage[T1]{fontenc}    
\usepackage[hidelinks]{hyperref}       
\usepackage{url}            
\usepackage{booktabs}       
\usepackage{amsfonts,amsmath}       
\usepackage{nicefrac}       
\usepackage{microtype}      
\usepackage{lipsum}
\usepackage{graphicx}
\usepackage{bm}
\graphicspath{ {./images/} }

\newcommand\blfootnote[1]{%
  \begingroup
  \renewcommand\thefootnote{}\footnote{#1}%
  \addtocounter{footnote}{-1}%
  \endgroup
}

\title{Aerodynamic force reconstruction using physics-informed Gaussian processes}

\author{
 Gledson Rodrigo Tondo$^{\star}$ \\
  Bauhaus-Universität Weimar\\
  Weimar, Germany \\
  \And
 Igor Kavrakov \\
  University of Cambridge\\
  Cambridge, United Kingdom \\
  \And
 Guido Morgenthal \\
  Bauhaus-Universität Weimar\\
  Weimar, Germany \\
}

\begin{document}
\maketitle
\begin{abstract}
Accurate modeling of aerodynamic loads is essential for understanding and predicting the responses of complex structural systems. However, these models often rely on simplifications of the true physical forces, introducing assumptions that can limit their accuracy. Validating such models becomes particularly challenging in the presence of noisy or incomplete data. To address this, we introduce a probabilistic physics-informed machine learning approach designed to reconstruct the underlying aerodynamic loads from noisy measurements of structural dynamic responses. The model avoids overfitting, eliminates the need for regularization schemes, and allows for the use of heterogeneous and multi-fidelity data during the training process. The efficacy of the approach is demonstrated through the reconstruction of aerodynamic loads on the Great Belt East Bridge, simulated under a linear unsteady assumption. Results show a strong agreement between true and predicted loads, particularly related to root mean squared errors, magnitude, phase angle and peak values of the signals. The method for load reconstructing holds broad applicability, such as modeling validation, future load estimation, and structural damage prognosis.
\end{abstract}


\section{Introduction}\blfootnote{ICWE 16 - International Conference of Wind Engineering\\ \hspace*{5mm} Florence, IT, 2023\\
\hspace*{5mm} \url{https://doi.org/10.1007/978-3-032-15130-8_20}}

\label{sec:1}

Aerodynamics is a critical field that plays a crucial role in the design and performance of long-span bridges. One of the main challenges in aerodynamics is the accurate prediction of the forces acting on a system, and while several models for this exist, they are generally based on different assumptions and simplifications of the true physical behavior \cite{Kavrakov2018}. 

In contrast, back-calculating the forces that generated a dynamic response is a complex task that has been widely studied~\cite{Sanchez2014}. Traditional methods for aerodynamic load reconstruction involve the construction of impulse response matrices~\cite{Amiri2017, Zhi2016, Chang2019} or the application of augmented Kalman filters~\cite{Zhi2018}, but rely on user-defined regularization parameters due to the noisy nature of measurement data. {The development of the latent force model~\cite{Alvarez2013} overcame the regularization problem by providing a linearized approach to generate physically inspired covariance functions for Gaussian process (GP), which has been successfully applied to reconstruct aerodynamic modal forces for the Hardanger bridge in Norway~\cite{Petersen2022}}. Similar to the Kalman filter, the method approximates the error covariance matrix at each discrete time step and, therefore, relies on a constant sampling interval and supports only one single data set at a time.

In this paper, a physics-informed machine learning approach for force reconstruction is presented, leveraging the knowledge of the physical system behavior given by mathematical formulations to construct covariance kernels for a Gaussian process regression model \cite{Tondo2023a,Tondo2025,Williams2006} (Sec.~\ref{sec:2}). The method provides an automatic trade-off balance between data fitting and model complexity, which naturally avoids overfitting and, due to its Bayesian nature, provides prediction uncertainty quantification. To demonstrate the proposed force reconstruction method, the linear unsteady model is employed to evaluate the structural response of the Great Belt East Bridge (Sec.~\ref{sec:gbanalysis}). The responses are then contaminated with white noise and used as input to the force reconstruction model, yielding in turn a stochastic model for the underlying aerodynamic force (Sec.~\ref{sec:forcerec}). Finally, a summary of the results and conclusions is provided in Sec.~\ref{sec:summary}.

\section{Physics informed Gaussian process for force reconstruction}
\label{sec:2}

{The dynamic response of a harmonic oscillator given an arbitrary time-varying load $F$ is given by the scalar equation}

\begin{equation}\label{eq:eq1}
	m \ddot{z} + 2m \zeta \omega_n \dot{z} + m \omega_n^2 z = F,
\end{equation}

\noindent where $m$ is the mass, $\zeta$ is the damping ratio, $\omega_n$ is the circular natural frequency and $z$, $\dot{z}$ and $\ddot{z}$ are the modal displacement, velocity and acceleration, respectively. The responses are measurable and, in general, contain different levels of noise based on the quality of the measurement system. {The displacement response, for instance, can be stochastically modeled as $z = f(t) + \epsilon$, where $t$ is time and $\epsilon \sim \mathcal{N} (0, \sigma^2_z)$ is the independent and identically distributed (i.i.d.) noise with variance $\sigma_z^2$, while the velocities and accelerations follow a similar principle. }

Assuming the displacement response is a sample from a zero-mean Gaussian process yields  

\begin{equation}\label{eq:eq3}
	z(t) \sim \mathcal{G \! \! P} \left( 0, k_{zz} (t,t') + \sigma_z^2 \delta (t,t') \right),
\end{equation}

\noindent where $\delta$ is the Kronecker-Delta, $t$ and $t'$ are time instants when deflections are measured and $k_{zz}$ is a covariance kernel that reflects the model assumptions. Because the displacement is assumed to be continuous and smooth in time, $k_{zz}$ is modeled herein by the squared exponential (SE) kernel, given by \cite{Williams2006}

\begin{equation}\label{eq:eq4}
	k_{zz}(t,t';\sigma_s,\ell) = \sigma^2_s \mathrm{exp} \left( -\frac{1}{2} \frac{(t-t')^2}{\ell^2} \right),
\end{equation}

\noindent where $\sigma_s^2$ is a variance scale and $\ell$ is the kernel's length scale, which controls the smoothness of the random functions.

\begin{figure}[!htb]
	\centering
	\includegraphics[scale=1]{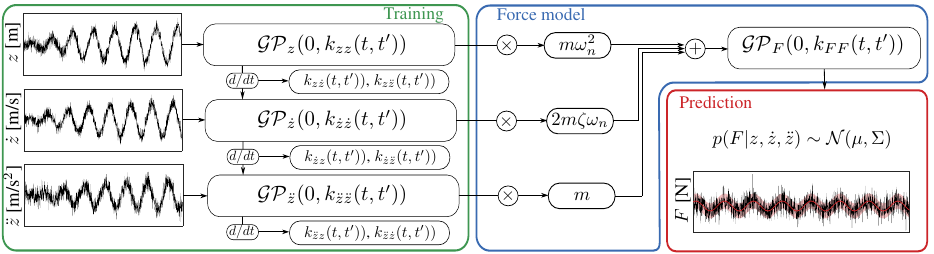}
	\caption{Framework for the physics-informed Gaussian process. Models for different data types are created and jointly trained (green). The force model is built using Eq. (\ref{eq:eq1}) (blue) and used for predictions (red).}
	\label{fig:figure1}
\end{figure}

Exploiting the linear time-derivative relation between displacements, velocities and accelerations, covariance kernels can be created to relate different physical measurement types (e.g. $k_{z \dot{z}} = dk_{zz}/dt'$ or $k_{\ddot{z} z} = d^2 k_{zz}/dt^2$), by applying the time derivative to the covariance kernel from Eq. (\ref{eq:eq4}). A summary of this procedure is shown in Fig. \ref{fig:figure1} (green block).

The SE kernel's parameters and the measurement noise represented by $\sigma_z$, $\sigma_{\dot{z}}$ and $\sigma_{\ddot{z}}$ are generally not known \textit{a priori}, and can be identified from data using maximum likelihood estimation, 

\begin{equation}\label{eq:eq5}
	\bm{\theta}_{\mathrm{opt}} = \mathrm{argmax}_{\bm{\theta}} \ \mathrm{log} p(\bm{y} | \bm{t}, \bm{\theta}) = \mathrm{argmax}_{\bm{\theta}} \left( -\frac{1}{2} \bm{y}^T \bm{K}^{-1} \bm{y} - \frac{1}{2} \mathrm{log} |\bm{K}| - \frac{N}{2} \mathrm{log} 2\pi \right),
\end{equation}

\noindent where $\bm{\theta} = \lbrace \sigma_s, \ell, \sigma_z, \sigma_{\dot{z}}$, $\sigma_{\ddot{z}} \rbrace$ is a vector containing the optimizable parameters, $\bm{y} = [\bm{z}, \dot{\bm{z}}, \ddot{\bm{z}}]^T \in \mathbb{R}^{N}$ contains the measurement training data at times $\bm{t} = [\bm{t}_z, \bm{t}_{\dot{z}}, \bm{t}_{\ddot{z}}]^T \in \mathbb{R}^{N}$, and $\bm{K}$ is a covariance matrix calculated as \cite{Tondo2023a}:

\begin{equation}\label{eq:eq6}
	\bm{K} = \begin{bmatrix}
		k_{zz}(\bm{t},\bm{t}') + \bm{I} \sigma_z^2  & k_{z\dot{z}}(\bm{t},\bm{t}') & k_{z\ddot{z}}(\bm{t},\bm{t}')\\
		
		k_{\dot{z} z}(\bm{t},\bm{t}') & k_{\dot{z} \dot{z}}(\bm{t},\bm{t}') + \bm{I} \sigma^2_{\dot{z}} & k_{\dot{z}\ddot{z}}(\bm{t},\bm{t}') \\
		
		k_{\ddot{z} z}(\bm{t},\bm{t}') & k_{\ddot{z}\dot{z}}(\bm{t},\bm{t}') & k_{\ddot{z}\ddot{z}}(\bm{t},\bm{t}') + \bm{I} \sigma^2_{\ddot{z}} \\
	\end{bmatrix}.
\end{equation}

The parameter identification depends, therefore, purely on the measurement data, meaning that no force information is used during the training stage of the physics-informed machine learning model. {In the identification, it is assumed that the noise in the measurement data is normally distributed, which may not be the case in realistic measurements. Extensions of the GP model to incorporate other types of noise are available in literature, but have not been considered as part of this study~\cite{Li2020}.} Additionally, although the covariance matrix from Eq. (\ref{eq:eq6}) uses displacements, velocities and accelerations, parameter identification can be carried out using only one, or multiple combinations of datasets, depending on their availability \cite{Tondo2023b}. 

With the identified parameters, a joint distribution can be built between the training data in $\bm{y}$ and the dynamic forces $\bm{F}$ (Fig.~\ref{fig:figure1}, blue block). Conditioning the forces on the data yields 

\begin{equation}\label{eq:eq7}
	p(\bm{F} | \bm{y}) = \mathcal{N}(\bm{\mu} = \bm{K}^T_* \bm{K}^{-1} \bm{y}, \bm{\Sigma} = \bm{K}_{**} - \bm{K}_*^T \bm{K}^{-1} \bm{K}_*), 
\end{equation}

\noindent where $\bm{K}_* = [k_{Fz} (\bm{t}_*, \bm{t}), k_{F\dot{z}} (\bm{t}_*, \bm{t}), k_{F\ddot{z}} (\bm{t}_*, \bm{t})]^T$, $\bm{K}_{**} = k_{FF} (\bm{t}_*, \bm{t}_*)$, $\bm{t}_*$ are the discrete time indexes for force prediction, and $k_{Fi}$ for $i \in \lbrace z, \dot{z}, \ddot{z}, F \rbrace$ are the kernels generated using the harmonic oscillator model given in Eq. (\ref{eq:eq1}), following the framework in Fig. \ref{fig:figure1} (red block).

\section{Aerodynamic analysis of the Great Belt east bridge}
\label{sec:gbanalysis}

\begin{figure}[!htb]
	\centering
	\includegraphics[width=0.9\textwidth,clip,trim={0 0.5mm 0 1.0mm},scale=1]{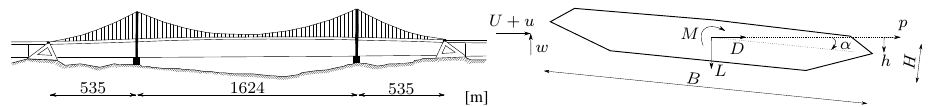}\\
	\includegraphics[width=0.9\textwidth]{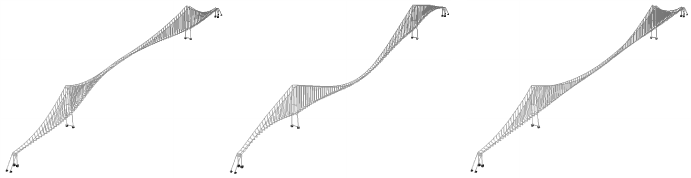}
	\caption{{The Great Belt East Bridge in Denmark. Top: elevation sketch (left) and deck coordinate system (right), with the mean wind speed $U$, turbulence components $u$ and $w$, degrees of freedom $p$, $h$ and $\alpha$, aerodynamic lift $L$, drag $D$ and moment $M$, and section width $B$ and height $H$. Bottom: the first sway ($f=0.052$~Hz), bending ($f=0.100$~Hz) and torsional ($f=0.278$~Hz) mode shapes.}}
	\label{fig:figure2}
\end{figure}

In order to evaluate the results of the dynamic force reconstruction method developed in Sec.~\ref{sec:2}, an aerodynamic analysis of a numerical model of the Great Belt East Bridge \cite{Larsen1993} in Denmark is carried out, yielding both the true applied forces and the correspondent responses. A sketch of the suspension bridge, along with its cross-section and the coordinate system for the aerodynamic forces, and the first three mode shapes in lateral sway, bending and torsional directions are shown in Fig.~\ref{fig:figure2}.

A reduced order model based on the linear unsteady (LU) \cite{Caracoglia2003,Kavrakov2017} assumption is used to numerically obtain the aerodynamic response of the bridge. Turbulent wind time-histories are generated using the von Karman spectrum, characterized by the turbulence intensities $I_u = 8\%$ and $I_w = 6\%$, and length scales $L_u = L_w = 60$~m. The mean component is assumed as $U = 30$~m/s, and the time histories are correlated in space along the length of the deck.

\begin{figure}[!htb]
	\centering
	\includegraphics[scale=1,trim={0 -1mm 0 0mm}]{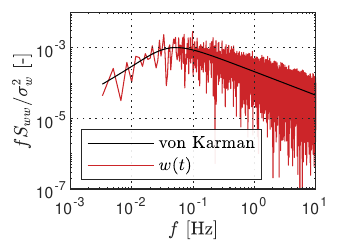}
	\includegraphics[scale=1,trim={0 -1mm 0 0mm}]{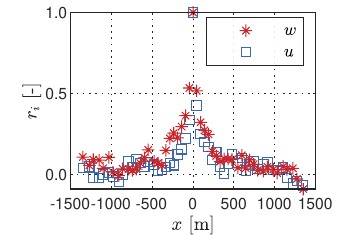}\\
	\includegraphics[scale=1,trim={0 1mm 0 1.0mm}]{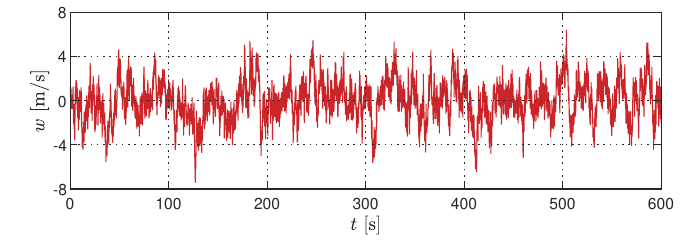}
	\caption{Top: turbulence spectrum in the vertical direction $w$ (left) and spatial correlation along the bridge length (right). Bottom: the vertical turbulence time history at midspan.}
	\label{fig:corrSpec}
\end{figure}

\begin{figure}[!htb]
	\centering
	\includegraphics[scale=1]{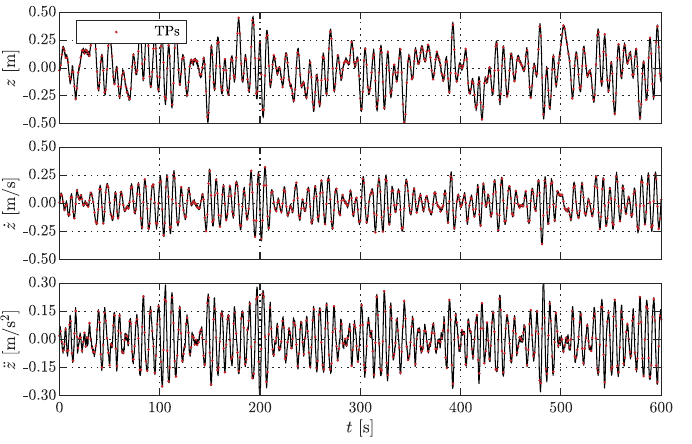}
	\caption{Modal responses for the first bending mode shape ($f=0.100$~Hz) and the selected training points (TPs). From top to bottom: modal displacements, velocities and accelerations.}
	\label{fig:modResp}
\end{figure}

The generated turbulence spectrum for the vertical component $w$ at midspan, and the wind signal correlation in space are shown in Fig.~\ref{fig:corrSpec}. A total of 22 modes are used in the dynamic analysis of the structure, which is carried out using a time interval of $\Delta t =$~0.05~s. The static wind coefficients and aerodynamic derivatives are obtained from~\cite{Kavrakov2018}, and buffeting analysis is carried out using Sears' admittance model. {The responses are obtained in the global coordinate system, synthetically contaminated with white noise with a signal-to-noise ratio SNR = 20, and further decomposed back into modal components to train the physics-informed machine learning model. This approach aims to simulate sensor readings, where measurement noise is always present and considered uncorrelated both in time and space.} {The noise intensity can negatively affect the quality of reconstructed forces, especially in stiffer modes of vibration (generally higher natural frequencies), where the mechanical admittance is in general low (see e.g.~\cite{Tondo2023a,Tondo2023b})}.  A sample of the modal responses for the first bending mode shape ($f=0.100$~Hz) are shown in Fig.~\ref{fig:modResp}.

\section{Dynamic force reconstruction}
\label{sec:forcerec}

Once the modal responses are obtained, a sparse selection of training points (TPs, cf. Fig.~\ref{fig:modResp}, top) is then selected at every $\Delta t_{\mathrm{train}} =$~1.25~s to train the GP model. This sparse selection of points is motivated by the fact that training a Gaussian process scales with $\mathcal{O}(N^3)$ complexity, with $N$ being the number of data points. Approximations of the inverse covariance matrix (e.g.~\cite{Snelson2007}) may accelerate training with the cost of adding prediction error, and are not investigated in this work. 

\begin{figure}[!htb]
	\centering
	\includegraphics[scale=1]{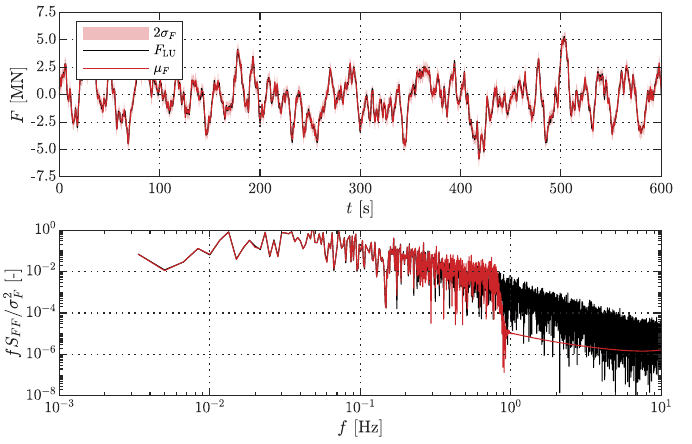}
	\caption{{True modal force signal $F(t)$ for the first structural mode, along with mean and 95\% confidence interval of predictions in time (top) and the correspondent power spectral density (bottom).}}
	\label{fig:LUForc}
\end{figure}

After identifying the optimal model parameters by maximum likelihood (c.f. Eq.~(\ref{eq:eq5})), the force model is conditioned on the training data yielding a stochastic representation of the dynamic forcing signal, according to Eq.~(\ref{eq:eq7}). {The mean prediction and confidence interval for the first bending mode shape ($f=0.100$~Hz) are shown in Fig.~\ref{fig:LUForc} (top), along with the true modal force computed with the linear unsteady model.}

The force predictions in the mean sense are in very good agreement with the true values calculated by the LU model. {The physics-informed GP outputs are a smoothed version of the original force.} In regions where high-frequency content dominates the forcing signal, the GP model captures the behavior as part of its uncertainty range. This is further observed in Fig.~\ref{fig:LUForc} (bottom), where the prediction power spectral density is compared with the original forcing signal. 

The regressed force captures the low-frequency content of the LU model force accurately, and the quality is inversely proportional to increasing frequencies. The predicted force has a very low spectral amplitude for frequencies higher than $f=1/\Delta t_{\mathrm{train}}=0.8$~Hz~, corresponding to the time interval of the training data. Increasing the time discretization of the training data, therefore, improves prediction quality and allows the model to capture high frequency content more accurately, with the cost of increased training time.

\begin{figure}[!htb]
	\centering
	\includegraphics[scale=1]{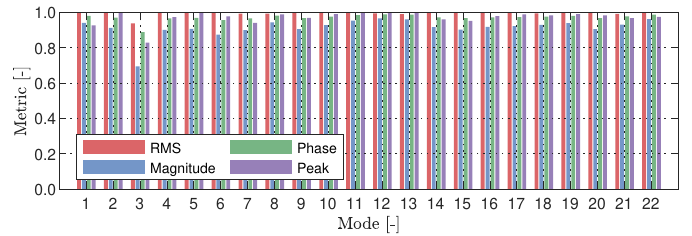}
	\caption{Comparison metric values between the original forces from the linear unsteady model and the mean values obtained from the physics-informed Gaussian process model.}
	\label{fig:CompMat}
\end{figure}

To evaluate the overall quality of the regressed modal forces, a comparison between the linear unsteady forces and the results obtained from physics-informed Gaussian process is carried out based on four metrics proposed by~\cite{Kavrakov2020}. The metrics evaluate the similarity of the signal's root mean square (RMS), localized magnitude, phase angle and absolute peak values in a normalized manner, such that values equal to unity mean perfect correspondence. The results are shown in Fig.~\ref{fig:CompMat}, and indicate a good agreement for nearly all modal components and comparison metrics.

\begin{figure}[!htb]
	\centering
	\includegraphics[scale=0.95]{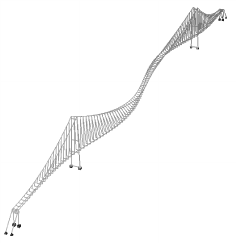} \hspace*{1cm}
	\includegraphics[scale=0.99]{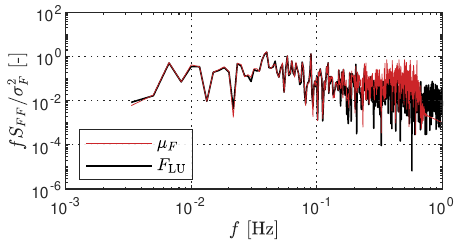}
	\caption{Left: the numerical model's third mode shape ($f=0.123$~Hz). Right: the PSD of the linear unsteady modal force and the correspondent physics-informed Gaussian process mean regression result.}
	\label{fig:Mode3}
\end{figure}

\begin{figure}[!htb]
	\centering
	\includegraphics[scale=1]{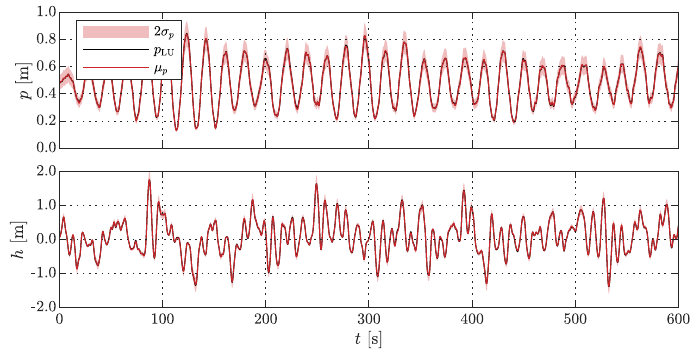}
	\caption{Global responses at midspan in drag direction ($p$, top) and lift direction ($h$, bottom) for the original LU model and computed with the GP regressed forces.}
	\label{fig:NodResp}
\end{figure}

The worst regression performance for all metrics was observed in the third mode shape ($f=0.123$~Hz), corresponding to a structural lateral sway, as shown in Fig.~\ref{fig:Mode3} (left). Analysis of the PSD of the regressed modal forces (c.f. Fig.~\ref{fig:Mode3}, right) indicates a good agreement in low frequency components ($f < 0.200$~Hz) of the forcing signal, and a significant increase of spectral amplitudes in the regressed force for frequencies above $0.200$~Hz in comparison to the linear unsteady model. This results from the addition of noise to the original signal in the global coordinate system, which has a higher impact in this particular mode due to its relative low contribution in the overall global responses. This is further shown in Fig.~\ref{fig:NodResp}, where the responses in the global coordinate system are shown for the LU model and compared with the ones calculated from the forces regressed from the physics-informed GP model.

\section{Summary and conclusions}
\label{sec:summary}

In summary, the constructed physics-informed machine learning model combines the strengths of machine learning methodologies with the structural behavior represented by mathematical models. This integration allows for the use of heterogeneous and multi-fidelity data during the training phase, and overcomes the need for regularization schemes. Moreover, the covariance kernels are analytically derived, yielding closed-form expressions by directly applying linear differential operators to a chosen base kernel, chosen here as the squared exponential one.

The model was employed to reconstruct the aerodynamic forces acting on the Great Belt East Bridge, as determined by the linear unsteady model. In order to replicate realistic measurement conditions, noise was introduced into the global structural responses. These responses were then transformed into their modal components and utilized to obtain the optimal model parameters via maximum likelihood. This approach balances data fitting and model complexity, avoiding overfitting issues and accommodating datasets with irregular sampling intervals for training purposes.

After training with measurement data, the models were applied to predict the underlying dynamic forces and the results were further compared with the original forces derived from the linear unsteady model. Despite the existence of intricate interactions between the wind field and resulting aerodynamic load, the model demonstrates a good ability to accurately regress the underlying force, incorporating high-frequency content and measurement noise into its uncertainty range. In scenarios where the structural motion exhibits minimal amplitude in comparison to the noise content of the signal, however, the results exhibit favorable agreement in low-frequency ranges. In contrast, in high-frequency ranges, the model tends to align with the noise inherent in the signal. 

In general, precise procedures for regressing dynamic forces serve as effective tools to investigate the accuracy and precision of models, particularly when applied to complex structures such as long-span bridges. The combination of the regressed forces with structural or load models can provide insights into the behavior of the structure and the characteristics of the wind field. This analysis gains further relevance in the context of climate change, considering its potential impacts on the wind field, which may deviate significantly from the assumptions made during the initial design phase. Additional potential applications of the developed model include also future load estimation and structural damage prognosis.

\bibliographystyle{unsrt}  
\bibliography{references}

\end{document}